\pdfoutput=1

\documentclass[11pt]{article}

\usepackage[preprint]{acl}

\usepackage{times}
\usepackage{latexsym}

\usepackage[T1]{fontenc}

\usepackage[utf8]{inputenc}

\usepackage{microtype}

\usepackage{inconsolata}

\usepackage{multirow}
\usepackage{graphicx}
\usepackage{arydshln}
\usepackage{subcaption}
\usepackage{amsmath}
\usepackage{amssymb}
\usepackage{microtype}
\usepackage{amsfonts}

\title{Representation Learning with Conditional Information Flow Maximization}

\author{Dou Hu$^{1,2}$
    \and Lingwei Wei$^{1}$\thanks{Corresponding author.}
    \and Wei Zhou$^{1}$
    \and Songlin Hu$^{1,2}$\footnotemark[1]
     \\
    $^{1}$ Institute of Information Engineering, Chinese Academy of Sciences \\
    $^{2}$ School of Cyber Security, University of Chinese Academy of Sciences  \\
    \texttt{\{hudou, weilingwei, zhouwei, husonglin\}@iie.ac.cn} \\
}

\begin{document}
\maketitle
\begin{abstract}
This paper proposes an information-theoretic representation learning framework, named conditional information flow maximization, to extract noise-invariant sufficient representations for the input data and target task. It promotes the learned representations have good feature uniformity and sufficient predictive ability, which can enhance the generalization of pre-trained language models (PLMs) for the target task. Firstly, an information flow maximization principle is proposed to learn more sufficient representations for the input and target by simultaneously maximizing both input-representation and representation-label mutual information. 
Unlike the information bottleneck, we handle the input-representation information in an opposite way to avoid the over-compression issue of latent representations. 
Besides, to mitigate the negative effect of potential redundant features from the input, we design a conditional information minimization principle to eliminate negative redundant features while preserve noise-invariant features. Experiments on 13 language understanding benchmarks demonstrate that our method effectively improves the performance of PLMs for classification and regression. Extensive experiments show that the learned representations are more sufficient, robust and transferable.

\end{abstract}

\section{Introduction}
The goal of deep representation learning \cite{DBLP:journals/nature/LeCunBH15} is to transform the raw observational data into low-dimensional representations that are essential for various downstream tasks. In recent years, information-theoretic representation learning has been widely studied, aiming to discover useful representations in a principled manner. The InfoMax principle \cite{DBLP:journals/computer/Linsker88} has extensive applications in the field of self-supervised representation learning \cite{DBLP:journals/corr/abs-1807-03748,DBLP:conf/iclr/HjelmFLGBTB19,DBLP:conf/iclr/TschannenDRGL20}. In supervised scenarios, minimizing the standard cross-entropy is actually equivalent to maximizing the mutual information between the representations and the target task
\cite{achille2018information,DBLP:conf/eccv/BoudiafRZGPPA20}. But InfoMax tends to preserve potential redundant features that are irrelevant to the given target, leading to biased representations. 
Another noteworthy line of information-theoretic research is built upon the information bottleneck (IB) principle \cite{tishby1999information,tishby2015deep}, which aims to discover compact and informative representations that can reduce redundant features from the inputs \cite{DBLP:conf/iclr/AlemiFD017,DBLP:conf/icml/BelghaziBROBHC18,DBLP:journals/entropy/Fischer20,DBLP:conf/cvpr/AnJC23,DBLP:conf/aaai/0001WLZH24}.
IB seeks to find a maximally compressed representation of the input that preserves as much information as possible about the target, striking a balance between compression and prediction.

However, in the information flow \cite{DBLP:conf/icml/GoldfeldBGMNKP19} of neural networks, directly reducing the mutual information between the input $X$ and representations $Z$ would violate the sufficiency constraint, and may lose the necessary information for the target task $Y$. Under the Markov chain constraint $Y \rightarrow X \rightarrow Z$, it's hard to determine beforehand how close we are to optimal compression, and this can easily lead to the over-compression issue of latent representations \cite{DBLP:journals/entropy/Fischer20,DBLP:conf/aaai/0001WLZH24}. As a result, current IB-based methods would yield insufficient representations for the target task, and hamper prediction ability of neural networks.

To ensure sufficiency for the target task and mitigate the negative effect of redundant features, we propose a principled representation learning framework, named conditional information flow maximization (CIFM), to extract noise-invariant sufficient representations for the input and target. It promotes the learned representations have good feature uniformity and sufficient predictive ability, which can enhance the generalization of pre-trained language models (PLMs) for the target task.

Firstly, we propose an information flow maximization (IFM) principle to learn more sufficient representations for the input and target. It simultaneously maximizes both input-representation and representation-label mutual information. Maximizing input-representation information $I(X;Z)$ ensures sufficiency of the representations for the input $X$ and preserves information relevant to the target $Y$, and maximizing representation-label information $I(Y;Z)$ captures necessary information relevant to the target $Y$. In this way, the learned representations can be of good uniformity and sufficient predictive ability. Unlike IB that minimizes input-representation information, we handle the information in an opposite way to avoid the over-compression issue of latent representations.

Besides, we design a conditional information minimization (CIM) principle to mitigate the negative effect of potential redundant features from the input. In the information flow of $X \rightarrow Z$, InfoMax may introduces excessive and noisy information \cite{DBLP:conf/nips/Tian0PKSI20}. For IFM, the task-irrelevant redundant nuisances features obtained by maximizing $I(X;Z)$ interfere with the optimization of maximizing $I(Y;Z)$. There are spurious correlations among these redundant features, forcing the model to learn a biased representation $Z$. As a conditional regularization term for IFM, the CIM principle eliminates negative redundant features while preserves noise-invariant features from inputs. Under the IFM principle with the conditional regularization, CIFM can extract noise-invariant sufficient representations for the input and target.

We conduct experiments on 13 natural language understanding benchmarks.
The results demonstrate that CIFM can significantly improve the performance of PLMs for classification and regression. Our CIFM framework consistently achieves the best average performance compared to other methods, including 4 universal models and 7 representative deep representation learning technologies under different backbone models. 
For instance, with the RoBERTa backbone, CIFM improves average performance by \textbf{+3.8\%} and \textbf{+1.9\%} for classification and regression tasks compared to CE/MSE, respectively.
Extended experiments prove that CIFM can enhance the model's generalization including out-of-distribution and data-constrained scenarios, robustness to random and adversarial noise, and transferability to new tasks. And the results also indicates that the learned representations by CIFM are more sufficient, robust and transferable.

The contributions are as follows:
    1) we propose an information flow maximization principle to learn more sufficient representations for the input and target by simultaneously maximizing both input-representation and representation-label information.  
    2) We design a conditional information minimization principle to eliminate negative redundant features while preserve noise-invariant features from the input.    
    3) We present an information-theoretic CIFM framework to learn noise-invariant sufficient representations for the input and target. It can enhance the generalization of PLMs for better language understanding. 
    4) Experiments on 13 language understanding benchmarks demonstrate that CIFM achieves better performance under different backbone models. Extensive experiments show that the learned representations are more sufficient, robust and transferable.\footnote{The source code is available at \url{https://github.com/zerohd4869/CIFM}}

\begin{figure*}[t]
    \centering
    \includegraphics[width=0.84\linewidth]{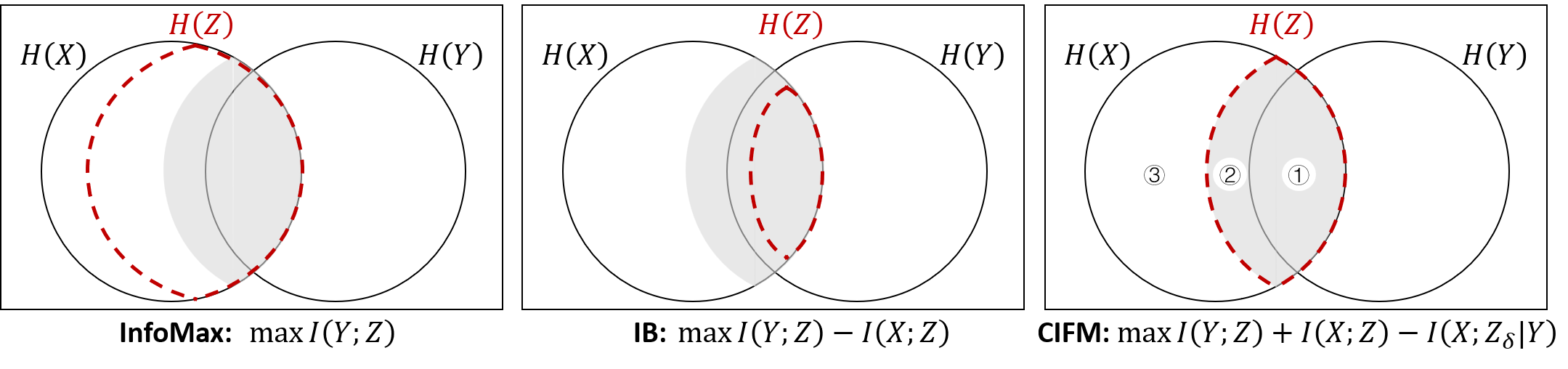}
    \caption{Venn information diagram comparison of our CIFM with existing principles. The learned representations by each principle is circled by the red dashed line.
    }
    \label{fig:model}
\end{figure*}

\section{Methodology}
This section presents a new information-theoretic representation learning framework, named {conditional information flow maximization} ({CIFM}), to extract sufficient representations for the input data and target task, as well as eliminates negative redundant features from the input. It contains two principles including information flow maximization and conditional information minimization.

\subsection{Information Flow Maximization}
The IB principle can reduce redundant features from the inputs by finding a compressed representation of the input that maximally preserves information about the output. However, in the information flow \cite{DBLP:conf/icml/GoldfeldBGMNKP19} of neural networks, directly minimizing the mutual information between the input $X$ and representations $Z$ would violate the sufficiency constraint, and may lose the necessary information for the target task $Y$.
\citet{DBLP:journals/entropy/Fischer20} and \citet{DBLP:conf/aaai/0001WLZH24} have shown that the compression term $I(X;Z)$ would reduce necessary information related to target task $Y$ under the Markov chain constraint $Y \rightarrow X \rightarrow Z$. As a result, current IB-based methods would yield insufficient representations for the target task, and hamper prediction ability of neural networks.

As shown in Figure~\ref{fig:model}, regions being minimized by $I(X;Z)$ overlap with relevancy regions being maximized by $I(Y;Z)$. 
To preserve informative signals from the input $X$ relevant to the target $Y$ (i.e., region \textcircled{1}) as much as possible in the information flow of $X \rightarrow Z$, we use the idea of InfoMax \cite{DBLP:journals/computer/Linsker88} to maximize $I(X;Z)$.
Besides, we also maximize $I(Y;Z)$ to capture necessary information relevant to the target task $Y$. Combing the above two terms, the principle of information flow maximization (IFM) simultaneously maximizes both input-representation and representation-label mutual information. It can be formulated as the maximization of the following Lagrangian,
\begin{equation}
    \max I(Y;Z) + \beta I(X;Z), 
    \label{eq:overall0}
\end{equation}
subject to the Markov chain constraint, i.e., $Y \rightarrow X \rightarrow Z$. $\beta$ is a parameter that balances the trade-off between the informativeness for $X$ and $Y$. 

In Equation~(\ref{eq:overall0}), the second term that maximizing input-representation information $I(X;Z)$ ensures sufficiency of the representations for the input $X$ and preserves informative signals from $X$ relevant to the target $Y$. The first term that maximizing representation-label information $I(Y;Z)$ captures necessary information relevant to the target $Y$. In this way, the learned representation by IFM can be of good uniformity and sufficient predictive ability. Unlike the IB principle that minimizes input-representation information, we handle the information in an opposite way to avoid the over-compression issue of latent representations.

\paragraph{Implementation of IFM}
The implementation of IFM contains two terms, i.e., maximizing $I(Y;Z)$ and $I(X;Z)$. 
First, we maximize the lower bound of $I(Y;Z)$ 
by estimating the conditional entropy of the target $Y$ given representations $Z$.
Following \citet{DBLP:journals/entropy/KolchinskyTW19} and \citet{DBLP:conf/aaai/0001WLZH24}, we use cross-entropy (CE) and mean squared error (MSE) as the estimators for classification and regression, respectively, i.e., $-\log \text{Softmax}(z_i,y_i)$ and $||z_i-y_i||^2$ for the $i$-th sample. Then, maximizing $I(X;Z)$ can be optimized by using the mutual information estimators, e.g., InfoNCE\footnote{InfoNCE is a softmax-based version of noise-contrastive estimation \cite{DBLP:journals/jmlr/GutmannH10}.} \cite{DBLP:journals/corr/abs-1807-03748} and MINE (mutual information neural estimator) \cite{DBLP:conf/icml/BelghaziBROBHC18}.

Here, we take InfoNCE as the default estimator for maximizing $I(X;Z)$. 
According to the information flow of  $Z \leftarrow X \rightarrow Z^\prime$, the Markov chain rule states that $I(X; Z) \geq I(Z; Z^\prime)$.
Maximizing the lower bound of $I(X;Z)$, i.e., $I(Z;Z^\prime)$, can be optimized by using InfoNCE estimator.
Specifically, for a sample $x_i$ with the representation $z_i$, its positive key is the augmented sample obtained by dropout, and its negative keys are the other samples in the batch of $N$ examples. In this case, we have,
\begin{equation}
\resizebox{1.0\linewidth}{!}{$
\begin{aligned}
    I(X;Z) &\geq I_{}(Z;Z^\prime) \\
    &\approx \log(K+1) + \frac{1}{N} \sum_{i=1}^{N} \log \frac{\exp(s(z_i, z_i^+) / \tau)}{\exp(s(z_i, z_i^+) / \tau) + \sum_{j=1}^{K} \exp(s(z_i,z_j^-) / \tau)},
    \label{eq:infonce}
\end{aligned}
$}
\end{equation}
where 
$z_i^+$ is the positive key and $z_j^-$ is the negative key.
$K$ represents the number of negative keys, which is $N-1$ in the default mode.
$s(\cdot)$ is a pairwise similarity function, i.e., cosine similarity, which can be seen as the dot product with $L_2$ normalization.
$\tau > 0$ is a scalar temperature parameter that controls the concentration or separation of probability distribution.

Alternatively, we also present how to utilize the MINE estimator to maximize $I(X;Z)$. Specifically, we use a neural network to approximate the lower bound of $I(X;Z)$, i.e.,
\begin{equation}
\resizebox{0.89\linewidth}{!}{$
\begin{aligned}
    I(X;Z) &\geq I_{\Omega}(X;Z) \\
    &= \sup_{\omega\in\Omega}\mathbb{E}_{\mathbb{P}_{{X}{Z}}}[T_{\omega}(Z|X)] - \log (\mathbb{E}_{\mathbb{P}_{X}\otimes\mathbb{P}_{Z}}[e^{T_{\omega}(Z|\tilde{X})}]),
      \label{eq:mine}
\end{aligned}
$}
\end{equation}
where ${\mathbb{P}_{{X}{Z}}}$ is the joint distribution and ${\mathbb{P}_{X}\otimes\mathbb{P}_{Z}}$  is the product of the marginals. 
$T_{\omega}$ is a function parametrized by an MLP (i.e., a fully-connected neural network with two hidden layers) with the parameter $\omega \in \Omega$. 
The expectations of $I_{\Omega}(X;Z)$ are estimated by shuffling the samples from the joint distribution along the batch axis. Additionally, an exponential moving average is used to make an unbiased estimate.

\subsection{Conditional Information Minimization}

In the information flow of $X \rightarrow Z$, InfoMax principle may introduce excessive and noisy information \cite{DBLP:conf/nips/Tian0PKSI20}. In our IFM, the task-irrelevant redundant nuisances features obtained by maximizing $I(X;Z)$ interfere with the optimization of maximizing $I(Y;Z)$. There are spurious correlations among these redundant features, forcing the model to learn a biased representation $Z$. 

To address the issue, we design a new conditional information minimization (CIM) term to eliminate negative redundant features while preserve noise-invariant features from inputs. As shown in Figure~\ref{fig:model}, we categorize all redundant information (i.e., region \textcircled{2} and \textcircled{3}) into two broad categories: negative (i.e., region \textcircled{3}) and non-negative (i.e., region \textcircled{2}), depending on whether it impedes the prediction ability of the target task. The negative features mean the nuisances that exist spurious correlations among redundant information and have a negative impact on task prediction. These non-negative features may encompass some inherent structured characteristics in the data, and preserving them during the learning process contributes to improving the model's generalization.

Given the target $Y$, we define the conditional effective mutual information between the input $X$ and the representations $Z$ as the subset of effective information that is irrelevant to the target $Y$.
Here, effective information  \cite{achille2019information,DBLP:conf/aaai/TerziAMS21} in the activations $Z_\delta$ can be seen as the amount of information about the input $X$ that is retained after adding the noise $\delta$. 
The CIM principle can be formulated as the minimization of the conditional effective information between the input X and the activations $Z_\delta$ given the target $Y$, i.e.,
\begin{equation}
    \min  I(X;Z_{\delta}|Y),
\end{equation}
where $Z_{\delta} = \phi_{w+\delta}(X)$ are the activations of the intermediate layer computed by the perturbed weights $w+\delta$.
Under the target $Y$, the higher the noise level $\delta$ (${\parallel \delta \parallel}_2 \le \epsilon$), the effective information in the activations $Z_{\delta}$ about $X$ gradually decreases to the optimum (i.e., region \textcircled{2} + \textcircled{3} in Figure~\ref{fig:model}) with the maximum noise level $\epsilon$.

\paragraph{Implementation of CIM}

Based on the relation between Fisher information and effective information \cite{achille2019information},
we use the Fisher Information of the weights to control the lower bound of the conditional effective information:
\begin{equation}
    \label{eq:shannon-fisher}
    \resizebox{0.89\linewidth}{!}{$
\begin{aligned}
    I(X;Z_{\delta}|Y) 
    \approx H(X|Y) - \mathbb{E}_{x,y} \left[\frac{1}{2} \log \left(\frac{(2\pi e)^k}{|F_{z|x,y}|}\right)\right], 
    \end{aligned}
    $}
\end{equation}
where $F_{z|x,y}$ represents the the Fisher information in the activations $Z_\delta$ given the input $X$ and target $Y$, i.e.,
$ F_{z|x,y} = \mathbb{E}_{z} [\nabla^2_{x} \log p(z|x,y)]$. $H(X|Y)$ is the conditional entropy of input $X$ given target $Y$. To reduce the conditional effective information between inputs $X$ and activations $Z_\delta$ given the target $Y$, it is sufficient to decrease the Fisher $|F_{z|x,y}|$, that is, increasing $\epsilon$ while keeping the target Y of the noise features remains the same. 

Inspired by \citet{DBLP:conf/aaai/TerziAMS21}, we use gradient-based adversarial training \cite{DBLP:journals/corr/GoodfellowSS14,DBLP:conf/iclr/MiyatoDG17}  to approximately estimate the minimization of the Fisher information $F_{z|x,y}$ in the activations $Z_\delta$ given the input $X$ and target $Y$. Formally, denote $(x, y)$ as a mini-batch input sampled from distribution $D$ and $p(y|x; {\theta})$ as a model with the parameter $\theta$. The worst-case perturbation $\delta$ under the given $Y$ can be computed by the back-propagation in the network, i.e., $\max_{\|\delta\|_q\leq\epsilon}{{KL}(p(\cdot|x;{\theta_\delta});p(\cdot|x;{\theta}))}$ with a Lagrange constraint $\mathcal{L}_{IFM}(x, y; \theta_\delta)$ and an $L_q$ norm constraint.
At each step of training, we identify the adversarial perturbations $\delta$ against the current network with the parameter $\hat{\theta}$, and put them on the  weights of first several hidden layers of the network. When using a Transformer-based network (e.g., RoBERTa) as the model backbone, we probabilistically put the perturbations on the weights of an embedding layer and the first hidden layer of the Transformer based on a predefined perturbation rate. With a linear approximation \cite{DBLP:journals/corr/GoodfellowSS14}, an $L_q$ norm-ball constraint, a certain radius $\epsilon$ for $\delta$,
and the original objective $\mathcal{L}_{IFM}$, the formulation of adversarial objective $\mathcal{L}_{CIM}$ is,
\begin{equation}
\resizebox{0.84\linewidth}{!}{$
\begin{aligned}
& \min_{\theta}\mathbb{E}_{{(x,y)}\sim{D}}\max_{\| \delta\|_2\leq\epsilon} \mathcal{L}_{IFM}(x, y; \theta_\delta),  \\
& \text{ where }
\delta = - \epsilon {g} / {\| g \|_q},
\text{ } g = \nabla_x \log p(y|x; \hat{\theta}).   
\end{aligned}
$}
\end{equation}
Under the objective of $\mathcal{L}_{IFM}$, the conditional information minimization loss $\mathcal{L}_{CIM}$ can be viewed as an adversarial regularizer.

\subsection{CIFM Framework}
We incorporate the conditional information minimization into information flow maximization, named conditional information flow maximization (CIFM).
The total optimization principle can be,
\begin{equation}
    \max I(Y;Z) + \beta I(X;Z) - I(X;Z_{\delta}|Y).
\end{equation}
The first two terms simultaneously maximize both input-representation and representation-label mutual information, which can learn more sufficient representations for the input and target. The last term eliminates negative redundant features while preserves noise-invariant features from the input to mitigate the negative effect of potential redundant features. Totally, CIFM can extract noise-invariant sufficient representations for the input and target. It promotes the learned representations have good uniformity and sufficient predictive ability, which can enhance the generalization of pre-trained language models for the target task.

As shown in Figure~\ref{fig:model}, given the target task, InfoMax maximizes information between the representations $Z$ and the target task $Y$ (e.g., the CE loss for classification), which preserves a lot of potential redundant features for the target task. 
There are spurious correlations among these redundant features (i.e., region \textcircled{3}) , forcing the model to learn biased representations $Z$. IB is prone to learning over-compressed representations due to the challenge in balancing between compression and task prediction. Directly minimizing the mutual information between the input $X$ and representations $Z$ can easily reduce necessary information  (i.e., region \textcircled{1}) for the target task $Y$ to some extent under the Markov chain constraint. 
Different from them, CIFM finds more sufficient representations (i.e., region \textcircled{1} and \textcircled{2}) for the input and target, as well as adversarially eliminates negative redundant features (i.e., region \textcircled{3}) from the input.

\section{Experiments}

\subsection{Experimental Setups}

\paragraph{Downstream Tasks and Datasets}
We conduct experiments on a variety of classification and regression tasks. Specifically, we utilize 10 classification benchmarks:
\textit{EmojiEval} \cite{DBLP:conf/semeval/BarbieriCRABBPS18} for emoji prediction, 
\textit{HatEval} \cite{DBLP:conf/semeval/BasileBFNPPRS19} for hate speech detection, 
\textit{IronyEval} \cite{DBLP:conf/semeval/HeeLH18} for irony detection,
\textit{OffensEval} \cite{DBLP:conf/semeval/ZampieriMNRFK19} for offensive language detection,
\textit{SentiEval} \cite{DBLP:conf/semeval/RosenthalFN17} for sentiment analysis,
\textit{StanceEval} \cite{DBLP:conf/semeval/MohammadKSZC16} for stance detection,
and 4 emotion-related benchmarks from different domains (i.e., \textit{EmotionEval} \cite{DBLP:conf/semeval/MohammadBSK18},
\textit{ISEAR} 
\cite{scherer1994evidence},
\textit{MELD} \cite{DBLP:conf/acl/PoriaHMNCM19},
and \textit{GoEmotions} \cite{DBLP:conf/acl/DemszkyMKCNR20}) for categorical emotion analysis. 
For regression, we use 3 benchmarks:
\textit{STS-B} \cite{DBLP:conf/semeval/CerDALS17} for semantic similarity prediction,
\textit{CLAIRE} \cite{DBLP:conf/semeval/RothAS22} for plausible clarification ranking,
and \textit{EmoBank} \cite{DBLP:conf/eacl/HahnB17} for dimensional emotion analysis.
More descriptions can be found in Appendix~\ref{sec:app:data}.

\begin{table*}[t]
\centering
\small
\tabcolsep=1.5pt
\resizebox{.96\linewidth}{!}{$
\begin{tabular}{l|cccccccccc|c}
\hline
\multicolumn{1}{c|}{\multirow{2}{*}{{Methods}}} 
& \multicolumn{1}{c}{{EmojiEval}} 
& \multicolumn{1}{c}{{EmotionEval}} 
& \multicolumn{1}{c}{{HatEval}} 
& \multicolumn{1}{c}{{IronyEval}} 
& \multicolumn{1}{c}{{OffensEval}} 
& \multicolumn{1}{c}{{SentiEval}} 
& \multicolumn{1}{c}{{StanceEval}} 
& \multicolumn{1}{c}{{ISEAR}} 
& \multicolumn{1}{c}{{MELD}} 
& \multicolumn{1}{c}{{GoEmotions}} 
& \multicolumn{1}{|c}{\multirow{2}{*}{\textbf{Avg.}}} \\  
& M-F1 & M-F1 & M-F1 & F1(i.) &  M-F1  & M-Recall & M-F1 (a. \& f.) & M-F1 & M-F1 & M-F1 \\
\hline 
\multicolumn{1}{l|}{SVM$^\dag$} & 29.30 & 64.70 & 36.70 & 61.70 & 52.30 & 62.90 & 67.30 & - & - & - & - \\ 
\multicolumn{1}{l|}{FastText$^\dag$} & 25.80 & 65.20 & 50.60 & 63.10 & 73.40 & 62.90 & 65.40 & - & - & - & - \\ 
\multicolumn{1}{l|}{BiLSTM$^\dag$} & 24.70 & 66.00 & 52.60 & 62.80 & 71.70 & 58.30 & 59.40 & - & - & - & - \\ 
\multicolumn{1}{l|}{GPT-3.5} & {6.34}\tiny{$\pm$0.01} & {73.23}\tiny{$\pm$0.18} &  {48.30}\tiny{$\pm$0.11} & \textbf{66.81}\tiny{$\pm$3.26} & {63.71}\tiny{$\pm$0.13}  & {40.40}\tiny{$\pm$3.13}  & {39.45}\tiny{$\pm$0.10}  & {67.22}\tiny{$\pm$0.09}  & {41.46}\tiny{$\pm$0.11}  & {25.21}\tiny{$\pm$0.08}  & {47.21} \\ 
\hline
\multicolumn{12}{l}{\multirow{1}{*}{\textit{BERT backbone}}} \\  \hline 
\multicolumn{1}{l|}{CE} &  
{22.30}\tiny{$\pm$0.60} & {76.05}\tiny{$\pm$1.41} & {44.67}\tiny{$\pm$1.78} & {59.38}\tiny{$\pm$3.01} & {80.16}\tiny{$\pm$1.26} & {70.54}\tiny{$\pm$0.44} & {65.21}\tiny{$\pm$0.71} & {67.17}\tiny{$\pm$0.78} & {39.80}\tiny{$\pm$0.84} & {46.29}\tiny{$\pm$0.79} & 57.16 \\  
\multicolumn{1}{l|}{CE+CP} & 21.91\tiny{$\pm$0.71} & 76.28\tiny{$\pm$1.20} & 45.97\tiny{$\pm$2.93} & 64.06\tiny{$\pm$2.41} & 78.99\tiny{$\pm$1.57} & 70.68\tiny{$\pm$0.31} & 65.83\tiny{$\pm$0.39} & 67.20\tiny{$\pm$0.95} & 39.54\tiny{$\pm$1.61} & 46.39\tiny{$\pm$0.63} & 57.69
\\ 
\multicolumn{1}{l|}{CE+AT} & 
22.93\tiny{$\pm$0.70} & 75.08\tiny{$\pm$1.23} & 46.30\tiny{$\pm$3.61} & 64.23\tiny{$\pm$2.04} & 79.68\tiny{$\pm$1.59} & 70.55\tiny{$\pm$0.57} & 66.46\tiny{$\pm$1.13} & 65.70\tiny{$\pm$0.69} & 39.84\tiny{$\pm$0.38} & 47.37\tiny{$\pm$0.54} &  57.81     \\
\multicolumn{1}{l|}{CE+SCL} & 21.72\tiny{$\pm$0.51} & 75.43\tiny{$\pm$1.37} & 45.86\tiny{$\pm$1.15} & 65.39\tiny{$\pm$2.46} & 80.20\tiny{$\pm$0.56} & 70.70\tiny{$\pm$0.79} & 65.34\tiny{$\pm$0.60}  & 67.54\tiny{$\pm$0.64} & 40.00\tiny{$\pm$1.96} & 46.50\tiny{$\pm$0.46} & 57.87  \\  
\multicolumn{1}{l|}{VIB} &  21.31\tiny{$\pm$0.62} & 77.37\tiny{$\pm$0.71} & 45.99\tiny{$\pm$1.93} & 63.82\tiny{$\pm$1.00} & 80.37\tiny{$\pm$1.11} & 70.39\tiny{$\pm$0.31} & 65.43\tiny{$\pm$0.60} & 67.24\tiny{$\pm$0.57} & 38.52\tiny{$\pm$0.51} & 45.89\tiny{$\pm$1.10} & 57.63
\\
\multicolumn{1}{l|}{MINE-IB} & 
21.29\tiny{$\pm$0.31} & 76.60\tiny{$\pm$0.41} & 47.64\tiny{$\pm$2.11} & 65.86\tiny{$\pm$2.57} & 78.67\tiny{$\pm$2.28} & 69.85\tiny{$\pm$0.54} & 65.35\tiny{$\pm$0.88} & 67.62\tiny{$\pm$0.40} & 41.23\tiny{$\pm$0.67} & 46.87\tiny{$\pm$0.42} & 58.10
\\ 
\multicolumn{1}{l|}{MEIB} & 21.87\tiny{$\pm$0.73} & 76.70\tiny{$\pm$0.82} & 48.27\tiny{$\pm$1.72} & 65.87\tiny{$\pm$2.14} & 80.49\tiny{$\pm$0.81} & 
70.55\tiny{$\pm$0.57} & 65.59\tiny{$\pm$1.58} & 67.44\tiny{$\pm$0.50} & 39.30\tiny{$\pm$0.61} & 46.26\tiny{$\pm$0.81} & 58.23  
\\
\multicolumn{1}{l|}{\textbf{CIFM}} & 
{24.28}$^{*}$\tiny{$\pm$0.39} & {77.74}\tiny{$\pm$1.07} & {59.22}$^{*}$\tiny{$\pm$1.75} & 
{65.51}\tiny{$\pm$1.61} & 
{80.67}\tiny{$\pm$0.99} & 
{70.66}\tiny{$\pm$0.28} & {67.81}$^{*}$\tiny{$\pm$0.97} & {71.51}$^{*}$\tiny{$\pm$0.51} & {41.72}$^{*}$\tiny{$\pm$0.80} & {48.42}$^{*}$\tiny{$\pm$0.74} & {60.75} \\ 
\hline
\multicolumn{12}{l}{\multirow{1}{*}{\textit{RoBERTa backbone}}}  \\ \hline 
CE & 
30.25\tiny{$\pm$1.32} & 77.41\tiny{$\pm$1.33} & 45.49\tiny{$\pm$4.70} & 57.99\tiny{$\pm$4.96} & 78.74\tiny{$\pm$2.20} & 71.80\tiny{$\pm$0.93} & 66.78\tiny{$\pm$1.34} & 70.00\tiny{$\pm$0.45} & 39.23\tiny{$\pm$0.41} & 46.64\tiny{$\pm$1.15} & 58.43 \\  
CE+CP & 
31.12\tiny{$\pm$0.84} & 77.54\tiny{$\pm$0.70} & 48.59\tiny{$\pm$3.28} & 58.75\tiny{$\pm$6.19} & 79.50\tiny{$\pm$0.98} & 72.82\tiny{$\pm$0.29} & 66.89\tiny{$\pm$1.67} & 70.58\tiny{$\pm$0.71} & 40.74\tiny{$\pm$0.89} & 47.98\tiny{$\pm$0.65} & 59.45 \\  
\multicolumn{1}{l|}{CE+AT} & 
32.00\tiny{$\pm$0.93} & 77.30\tiny{$\pm$1.07} & 44.71\tiny{$\pm$4.76} & 60.17\tiny{$\pm$3.17} & 79.81\tiny{$\pm$1.11} & 72.51\tiny{$\pm$0.44} & 67.81\tiny{$\pm$0.95} & 70.97\tiny{$\pm$0.68} & 40.10\tiny{$\pm$0.60} & 47.89\tiny{$\pm$1.21} & 59.33   \\ 
CE+SCL & 
31.09\tiny{$\pm$1.85} & 76.98\tiny{$\pm$2.02} & 49.51\tiny{$\pm$2.86} & 60.71\tiny{$\pm$4.23} & 80.39\tiny{$\pm$0.83} & 
\textbf{73.16}\tiny{$\pm$0.44} 
& 66.73\tiny{$\pm$1.54} & 70.26\tiny{$\pm$0.45} & 40.64\tiny{$\pm$1.02} & 47.87\tiny{$\pm$0.86} & 59.72  \\    
VIB &   
{29.71}\tiny{$\pm$0.79} & {77.99}\tiny{$\pm$0.86} & {49.39}\tiny{$\pm$3.08} & {59.93}\tiny{$\pm$4.57} & {79.63}\tiny{$\pm$0.66} & {72.81}\tiny{$\pm$0.39} & {68.40}\tiny{$\pm$0.52} & {70.74}\tiny{$\pm$0.44} & {38.94}\tiny{$\pm$0.55} & {46.23}\tiny{$\pm$0.18}  & {59.38} 
\\
MINE-IB & 
31.70\tiny{$\pm$0.45} &  {78.79}\tiny{$\pm$0.58} & {46.39}\tiny{$\pm$2.82} & {57.39}\tiny{$\pm$8.27} & 79.76\tiny{$\pm$0.67} & 72.85\tiny{$\pm$0.56} & 67.27\tiny{$\pm$1.00} & 70.15\tiny{$\pm$0.58} & 41.80\tiny{$\pm$2.14} & 48.88\tiny{$\pm$1.04} & 59.50 \\ 
MEIB 
& {29.94}\tiny{$\pm$1.30} 
& 78.73\tiny{$\pm$0.90}  & 49.34\tiny{$\pm$2.42} 
& {60.54}\tiny{$\pm$2.70}  
& 79.68\tiny{$\pm$0.98} & 72.78\tiny{$\pm$0.29} & 67.89\tiny{$\pm$1.70} & 70.86\tiny{$\pm$0.61} 
& 39.00\tiny{$\pm$0.37}  
&
47.18\tiny{$\pm$1.15} & {59.59} \\   
\textbf{CIFM} & 
\textbf{32.32}\tiny{$\pm$0.87} & \textbf{79.63}$^*$\tiny{$\pm$0.57} & \textbf{59.55}$^*$\tiny{$\pm$0.91} & 
{63.02}$^*$\tiny{$\pm$3.81} 
& \textbf{80.96}$^*$\tiny{$\pm$0.55} & 72.93\tiny{$\pm$0.27} & \textbf{69.01}$^*$\tiny{$\pm$0.80} & \textbf{71.66}$^*$\tiny{$\pm$0.68} & \textbf{43.99}$^*$\tiny{$\pm$1.10} & \textbf{49.51}$^*$\tiny{$\pm$0.31} & \textbf{62.26} \\
\hline
\end{tabular}
$}
\caption{Classification evaluation (\%) on 10 benchmarks.  
BERT and RoBERTa are the backbone models for fine-tuning on each benchmark. $^\dag$ means the results are from \citet{DBLP:conf/emnlp/BarbieriCAN20}. For other methods, we run five random seeds and report the average result on test sets. Best results for each benchmark are highlighted in bold. $^{*}$ represents statistical significance over state-of-the-art scores for the same backbone under the $t$-test ($p < 0.05$).
}
\label{tab:classification}
\end{table*}

\begin{table*}[t]
\centering
\resizebox{0.74\linewidth}{!}{$
\begin{tabular}{l|ccccccc|c}
\hline 
\multicolumn{1}{c|}{\multirow{2}{*}{Methods}}    &  \multicolumn{2}{c}{STS-B} & \multicolumn{2}{c}{CLAIRE} & \multicolumn{3}{c|}{EmoBank} & \multirow{2}{*}{\bf Avg.}   \\  
& Spearman & Pearson & Spearman & Pearson & Pearson (v) & Pearson (a) & Pearson (d)  \\ 
\hline
MSE & {88.33}\tiny{$\pm$0.32} & {88.80}\tiny{$\pm$0.36} & 
{50.37}\tiny{$\pm$5.90} & {49.10}\tiny{$\pm$5.74} & {80.62}\tiny{$\pm$0.64} & {55.43}\tiny{$\pm$1.95} &	{49.51}\tiny{$\pm$1.64} & 66.72 \\
MSE+AT & 
{88.40}\tiny{$\pm$0.50} & {89.01}\tiny{$\pm$0.37} & {53.09}\tiny{$\pm$0.64} & {51.87}\tiny{$\pm$0.65} & {81.04}\tiny{$\pm$0.82}	& {56.73}\tiny{$\pm$1.16} &	{51.48}\tiny{$\pm$1.46} & 68.15 \\
VIB &  {88.45}\tiny{$\pm$0.50} & {89.01}\tiny{$\pm$0.40} & {52.86}\tiny{$\pm$0.88} & {51.66}\tiny{$\pm$0.78} & {79.41}\tiny{$\pm$0.82} & {55.07}\tiny{$\pm$0.96} &	{46.50}\tiny{$\pm$3.85} & 67.10 \\ 
MEIB & {88.61}\tiny{$\pm$0.14} & {89.13}\tiny{$\pm$0.17} & {52.85}\tiny{$\pm$0.72} & {51.39}\tiny{$\pm$0.81} & {79.32}\tiny{$\pm$0.63} & {55.58}\tiny{$\pm$1.15} &	{46.63}\tiny{$\pm$3.10} & 67.17 \\ 
\textbf{CIFM} &    \textbf{88.94}$^*$\tiny{$\pm$0.38} & \textbf{89.52}$^*$\tiny{$\pm$0.32} & \textbf{53.62}$^*$\tiny{$\pm$0.51} & \textbf{52.88}$^*$\tiny{$\pm$0.30} & \textbf{81.64}$^*$\tiny{$\pm$0.84} & 	\textbf{57.06}$^*$\tiny{$\pm$0.95} & 	\textbf{51.60}\tiny{$\pm$1.11} & \textbf{68.64} \\
\hline
\end{tabular}
$}
\caption{Regression evaluation (\%) on 3 benchmarks with RoBERTa backbone. $^{*}$ represents statistical significance over state-of-the-art scores under the $t$-test ($p < 0.05$).
}
\label{tab:Regression}
\end{table*}

\paragraph{Comparison Methods}

We compare against the 4 universal models (i.e., SVM \cite{cortes1995support}, FastText \cite{DBLP:conf/eacl/GraveMJB17}, BiLSTM \cite{hochreiter1997long}, and GPT-3.5\footnote{\url{https://chat.openai.com}}) and 7 representative deep representation learning technologies (i.e., 
CE/MSE, CE+CP \cite{DBLP:conf/iclr/PereyraTCKH17}, CE/MSE+AT \citep{DBLP:conf/iclr/MiyatoDG17}, CE+SCL \cite{gunel2020supervised}, VIB \cite{DBLP:conf/iclr/AlemiFD017,DBLP:conf/iclr/MahabadiBH21}, MINE-IB \cite{DBLP:conf/icml/BelghaziBROBHC18} and MEIB \cite{DBLP:conf/cvpr/AnJC23}) with 2 different backbone models. 
For these representation learning technologies, we use two PLMs, i.e., BERT \cite{DBLP:conf/naacl/DevlinCLT19} and RoBERTa \cite{DBLP:journals/corr/abs-1907-11692}, as the backbone models for fine-tuning on downstream tasks. Concretely, we use \textit{bert-base-uncased}\footnote{\url{https://huggingface.co/}\label{code}} and \textit{roberta-base}\textsuperscript{\ref{code}} to initialize BERT and RoBERTa for fine-tuning on downstream tasks. See Appendix~\ref{sec:comparison_method} for more details.

\paragraph{Evaluation Metrics} \label{sec:eval}
We use the same evaluation metric from the original tasks. For classification tasks, the macro-averaged F1 over all classes is applied in most cases. There are three exceptions: stance (macro-averaged of F1 of favor and against classes), irony (F1 of ironic class), and sentiment analysis (macro-averaged recall). For regression tasks, we use both Pearson and Spearman correlation coefficients on STS-B and CLAIRE, and Pearson correlation for each VAD dimension on EmoBank.
We also report a global metric based on the average of all dataset-specific metrics, i.e., $\frac{1}{T}\sum_{t=1}^T \frac{1}{N_t}\sum_{n-1}^{N_t} M_{t,n},$. Here, $M_{t,n}$ denotes the performance for the $n$-th metric in the $t$-th task, $N_t$ denotes the number of metrics in the $t$-th task, and $T$ refers to the number of tasks. Besides, for the target dataset, the $t$-test \cite{kim2015t} is used to verify the statistical significance of the differences between the results of our CIFM and the best non-CIFM methods using the same backbone.

\paragraph{Implementation Details}
All experiments are conducted on a single NVIDIA Tesla A100 80GB card. 
The validation sets are used to tune hyperparameters and choose the optimal model. For each method, we run five random seeds and report the average result of the test sets. The network parameters are optimized by using Adamax optimizer \citep{DBLP:journals/corr/KingmaB14} with the learning rate of $5e^{-5}$, the weight decay coefficient of $\{0, 0.01, 0.001\}$. More implementation details are listed in Appendix~\ref{sec:para}.

\begin{table}[t]
\centering
  \resizebox{0.98\linewidth}{!}{$
  \begin{tabular}{l|c|cc}
  \hline
  \multicolumn{1}{c|}{\multirow{2}{*}{Methods}} & \multicolumn{1}{c|}{\textbf{Avg. Classification}} & \multicolumn{1}{c}{\textbf{Avg. Regression}} \\
    & \multicolumn{1}{c|}{\it BERT / RoBERTa }  &  \multicolumn{1}{c}{\it RoBERTa }   \\
  \hline 
\textbf{CIFM} &  
\textbf{60.75} / \textbf{62.26} & \textbf{68.64} \\  
\ \ - w/o CIM & 58.96 / 60.96 & 66.77 \\
\ \ - w/o CIM \& IFM &  57.27 / 57.92 & 67.32 
\\
\hline
\end{tabular}
$}
  \caption{Ablation results (\%). We report the average results on 10 classification benchmarks and 3 regression benchmarks. See Appendix~\ref{sec:results_ablated} for the detailed results.
  }   \label{tab:abla}
\end{table}

\subsection{Overall Results} \label{sec:overall}
The overall results for 10 classification benchmarks and 3 regression benchmarks are summarized in Table~\ref{tab:classification} and Table~\ref{tab:Regression}, respectively. Our CIFM consistently obtains the best average performance over comparison methods. When using RoBERTa, CIFM improves average performance by  \textbf{+3.8}\% and \textbf{+1.9}\% for classification and regression tasks compared to CE/MSE, respectively. The results indicate the good generalization ability of our method to unseen test sets and show the superiority of our method. We notice that CIFM achieves big improvements for \textit{HatEval} and \textit{IronyEval}, i.e., \textbf{+14.1\%} macro-F1 scores and \textbf{+5.0\%} F1 scores of ironic. In \textit{HateEval}, there is a phenomenon of disparity in topic distribution between the validation set and the test set. Besides, the task of \textit{IronyEval} requires complex semantic understanding, and the differences between ironic and non-ironic texts are usually subtle. The results indicate that CIFM has a good generalization capability on the above scenarios, i.e., topic shifts and subtle semantic labels.

\begin{figure}[t]
    \centering
    \includegraphics[width=0.9\linewidth]{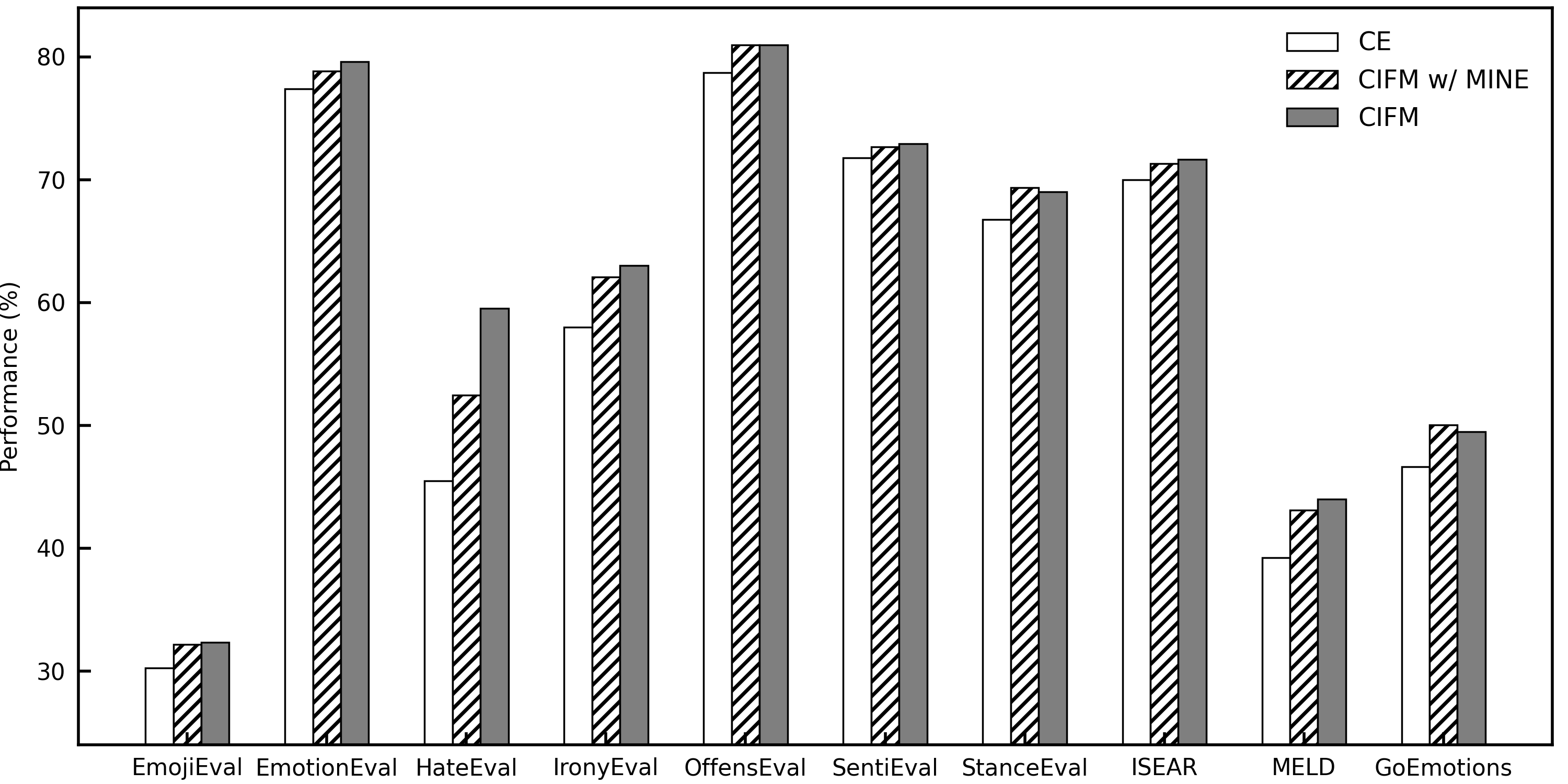}
    \caption{Comparison results of CIFM with different MI Estimators and the CE baseline on classification tasks. RoBERTa is the default backbone model.}
    \label{fig:cifm_imple}
\end{figure}

\begin{figure*}[t]
\centering
\includegraphics[width=0.9\textwidth]{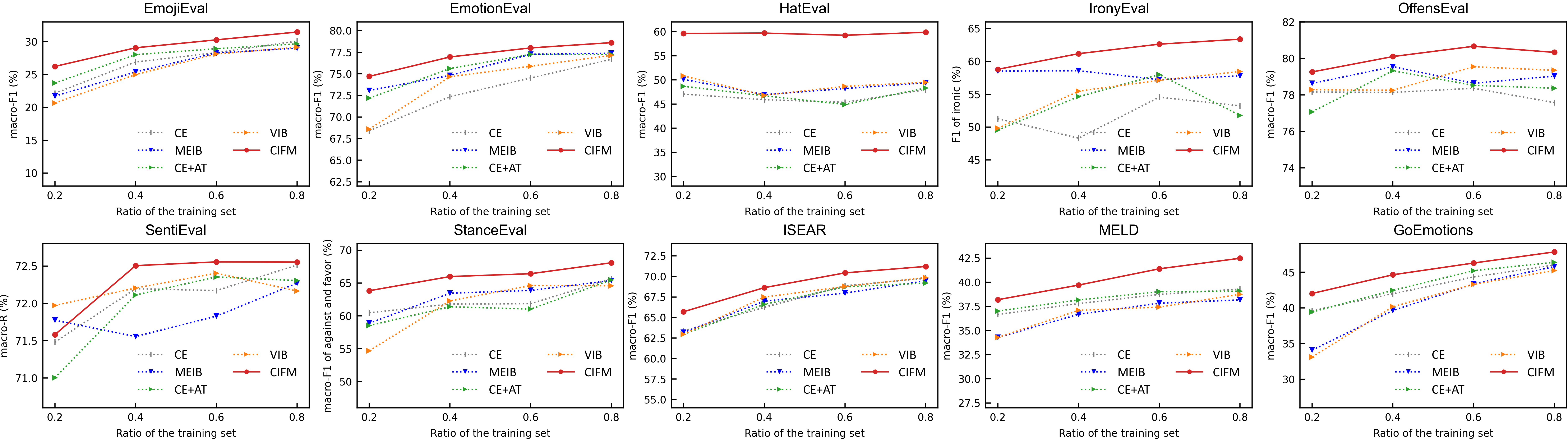} \caption{Results of different methods against different sizes of training set with RoBERTa backbone.}
\label{fig:training}
\end{figure*}

\begin{table}[t]
\centering
\resizebox{0.98\linewidth}{!}{$
\begin{tabular}{l|c|c|c|c}
\hline
\multicolumn{1}{c|}{\multirow{2}{*}{Methods}} 
& {EmotionEval}
& {ISEAR} 
& {MELD} 
&  \multirow{2}{*}{\bf Avg.}
\\ 
& $\rightarrow$   GoEmotions & 
 $\rightarrow$  GoEmotions & 
  $\rightarrow$  GoEmotions & \\ 
\hline 
CE      & 73.79\scriptsize{$\pm$2.57} & 42.99\scriptsize{$\pm$2.10} & 30.71\scriptsize{$\pm$0.54} & 49.16  \\
CE+AT   & 72.54\scriptsize{$\pm$3.89} & 44.11\scriptsize{$\pm$1.44} & 32.05\scriptsize{$\pm$1.69} & 49.57 \\ 
VIB     & 74.73\scriptsize{$\pm$3.52} & 41.88\scriptsize{$\pm$1.65} & 30.50\scriptsize{$\pm$1.05} & 49.03 \\
MEIB    & 75.55\scriptsize{$\pm$2.05} & 42.10\scriptsize{$\pm$0.61} & 30.11\scriptsize{$\pm$1.33} & 49.25 \\
\textbf{CIFM} & \textbf{75.88}\scriptsize{$\pm$0.93} & \textbf{58.85}\scriptsize{$\pm$1.93} & \textbf{36.44}\scriptsize{$\pm$1.76} & \textbf{57.06} 
\\
 \hline
\end{tabular}
$}
\caption{Out-of-distribution evaluation results (\%). For example, \textit{MELD} $\rightarrow$ \textit{GoEmotions} refers to training the model on training set of \textit{MELD} and predicting with the test set of \textit{GoEmotions}. We experiment with RoBERTa backbone. We run five random seeds and report the average results on test sets of target domains. Labels that do not appear in the training corpus are not evaluated. 
}
\label{tab:ood-performance}
\end{table}

\subsection{Ablation Study} \label{sec:abla}
\paragraph{Loss Analysis}
Table~\ref{tab:abla} shows an ablation study by removing the conditional information minimization (w/o CIM) and information flow maximization (w/o IFM). From results, the full CIFM achieves the best performance on classification and regression tasks. It proves the effectiveness of combing CIM and IFM. Specifically, learning more sufficient representations by IFM can boost the performance of classification tasks. Mitigating the negative effect of redundant features by CIM is more helpful for regression.

\paragraph{Effectiveness Evaluation with Different MI Estimators}
We further implement our CIFM by replacing the default InfoNCE estimator with MINE \cite{DBLP:conf/icml/BelghaziBROBHC18} to estimate $I(X,Z)$. Figure~\ref{fig:cifm_imple} shows comparison results of CIFM with different mutual information (MI) estimators and the CE baseline on the classification benchmarks. From results, our CIFM and its variant CIFM w/ MINE consistently achieve better results on all tasks. It confirms the effectiveness of our CIFM principle under different MI estimators.

\begin{figure*}[t]
\centering
\includegraphics[width=0.9\textwidth]{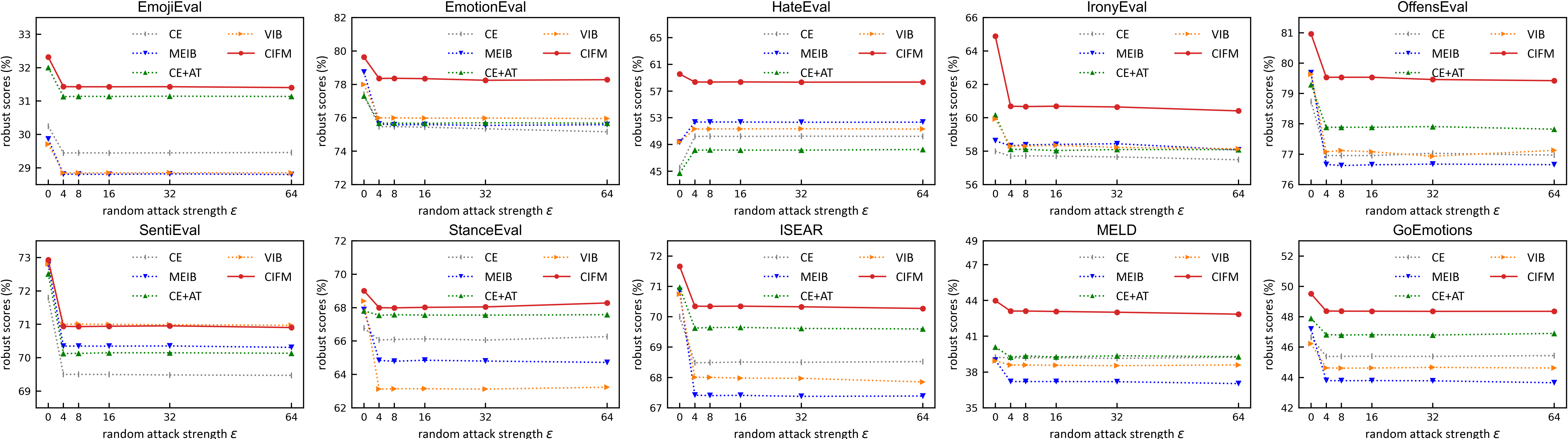}
\caption{
Robust scores against different random perturbation strengths. RoBERTa is the default backbone. 
}
\label{fig:robust_random}
\end{figure*}

\begin{figure*}[!ht]
    \centering
    \includegraphics[width=0.9\textwidth]{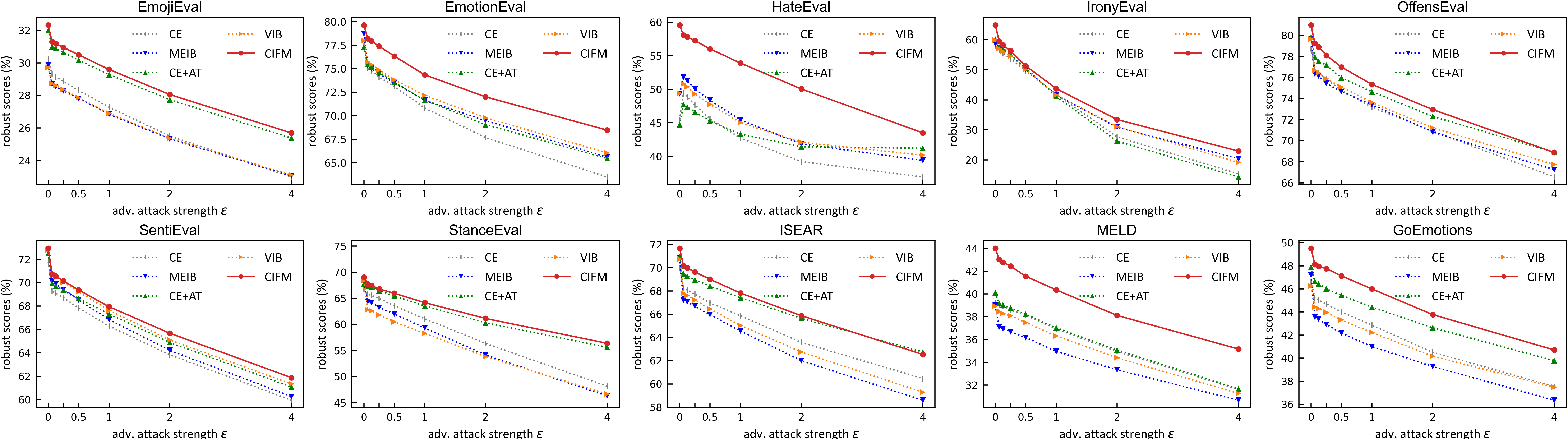} 
    \caption{Robust scores against different adversarial perturbation strengths. RoBERTa is the default backbone.}
    \label{fig:robust_adv}
\end{figure*}

\begin{table*}[t]
\centering
\resizebox{0.99\linewidth}{!}{$
\begin{tabular}{c|l|ccccccccc|c|c}
\hline 
\multicolumn{2}{c|}{\multirow{1}{*}{{Methods}}} & \multicolumn{10}{c|}{GoEmotions $\rightarrow$ Others} 
& \multicolumn{1}{c}{\multirow{1}{*}{SentiEval $\rightarrow$ Others}} \\ \hline
Cls. &  \multicolumn{1}{c|}{Pre-train Obj.} & \multicolumn{1}{c}{{EmojiEval}} 
& \multicolumn{1}{c}{{EmotionEval}} 
& \multicolumn{1}{c}{{HatEval}} 
& \multicolumn{1}{c}{{IronyEval}} 
& \multicolumn{1}{c}{{OffensEval}} 
& \multicolumn{1}{c}{{SentiEval}} 
& \multicolumn{1}{c}{{StanceEval}} 
& \multicolumn{1}{c}{{ISEAR}} 
& \multicolumn{1}{c|}{{MELD}} 
&  \multicolumn{1}{c|}{\multirow{1}{*}{\bf Avg.}}
& \multicolumn{1}{c}{\multirow{1}{*}{\bf Avg.}}\\ 
\hline 
\multirow{5}{*}{Linear}
& CE & 5.89 & 59.39 & 58.52 & \bf 49.36 & 74.86 & 66.42 & 38.47 & 49.58 & 31.57 & 48.23 
& 37.88 
\\ 
& CE+AT & 5.74 & 59.65 & 58.47 & 46.23 & 74.67 & 65.86 & 37.43 & 48.63 & 31.62 & 47.59
& 38.20 
\\ 
& VIB & 5.64 & 52.51 & 58.01 & 45.54 & 72.57 & 64.87 & \bf 39.05 & 47.11 & 30.00 & 46.14 
& 37.07 
\\ 
& MEIB & 4.44 & 51.10 & 57.64 & 44.67 & 71.38 & 62.79 & 37.89 & 47.98 & 28.61 & 45.17 
& 35.26
\\ 
& \textbf{CIFM} & \bf 8.10 & \bf 59.74 & \bf 59.33 & 47.96 & \bf 75.41 & \bf 67.03 & 38.41 & \bf 50.17 & \bf 31.71 & {\bf 48.65} 
& {\bf 40.44}
\\ 
\hline 
\multirow{5}{*}{CNN} 
& CE & 11.43 & 72.33 & 45.40 & \bf 59.35 & 75.84 & 70.22 & 56.89 & 61.88 & 41.29 & 54.96 
& 55.13
\\ 
& CE+AT & 10.74 & 73.03 & 49.16 & 57.36 & 78.30 & 70.49 & 55.29 & 61.93 & 41.20 & 55.28 
& 55.40 
\\ 
& VIB & 12.52 & 72.02 & 48.52 & 57.83 & 75.43 & 69.62 & 55.91 & 60.38 & 40.06 & 54.70 
& 53.07
\\ 
& MEIB & 10.03 & 71.45 & 47.09 & 51.39 & 75.10 & 68.43 & 47.33 & 58.93 & 38.55 & 52.03 
& 50.94
\\ 
& \textbf{CIFM} & \bf 15.19 & \bf 75.30 & \bf 50.58 & 57.14 & \bf 78.64 & \bf 70.99 & \bf 58.80 & \bf 64.71 & \bf 41.37 & {\bf 56.97} 
& {\bf 56.21}
\\ 
\hline
\end{tabular}$}
\caption{Transferability results (\%) of \textit{GoEmotions} $\rightarrow$ \textit{Others} and \textit{SentiEval} $\rightarrow$ \textit{Others} with RoBERTa backbone. \textit{GoEmotions} and \textit{SentiEval} are the source datasets for pre-training. Pre-train Obj. indicates the pre-training objective. Cls. refers to the classifier when fine-tuning. Best results for linear (i.e., Linear) and nonlinear (i.e., CNN) classifiers are highlighted in bold. Detailed results of \textit{SentiEval} $\rightarrow$ \textit{Others} are listed in Appendix~\ref{tab:transferability}.}
\label{tab:transfer_detailed1}
\end{table*}

\begin{figure*}[t]
\centering
\includegraphics[width=0.94\linewidth]{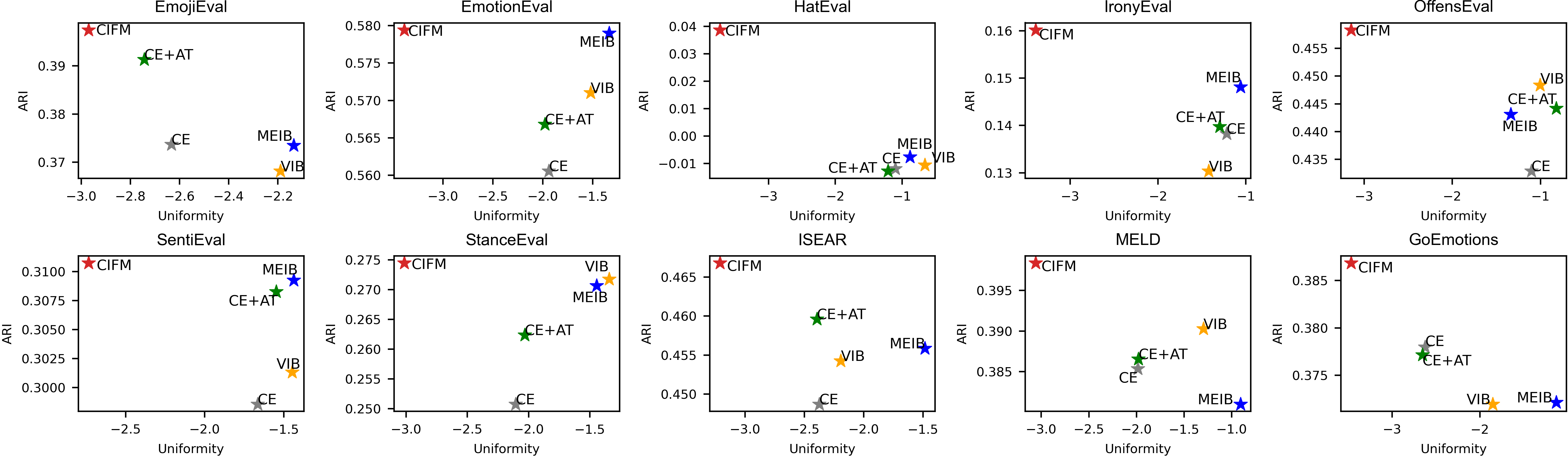}
\caption{
Quality analysis of the learned representations by different optimization objectives. 
The X-axis refer to the uniformity of hidden representations, and the Y-axis refer to the ARI score of output representations. The lower uniformity means the better sufficiency for the input, and the higher ARI means the better sufficiency for the target.
}
\label{fig:para}
\end{figure*}

\subsection{Generalization Evaluation} \label{sec:generalization}
We evaluate the generalization of CIFM under the following two settings, i.e., out-of-distribution (OOD) and data-constrained scenarios.

\paragraph{Out-of-Distribution Generalization Evaluation}
We choose emotion-related benchmarks (e.g., \textit{EmotionEval}, \textit{ISEAR}, \textit{MELD}, and \textit{GoEmotions}), which are collected from different domains and aim to predict the label of different emotion taxonomies. To implement OOD scenarios, we train the model on the original training set from a source domain, select the best model based on the validation set of the source domain, and test on the test set of a target domain. To avoid the interference of label mapping bias between different taxonomies, each model is trained on the dataset with coarse-grained taxonomy to predict the label for another dataset with fine-grained taxonomy (i.e., GoEmotions). Table~\ref{tab:ood-performance} shows the performance under OOD scenarios. CIFM obtains the best results on all OOD settings. Comparing to CE, CIFM achieves \textbf{+7.9}\% improvements in terms of average scores. This results exhibit that CIFM can promote the learning of more sufficient representations for target task than others, which enhances the generalization in handling OOD scenarios across different domain shifts.

\paragraph{Evaluation under Data-constrained Scenarios}
We experiment under different ratios of the training set to evaluate the generalization when training with data-constrained scenarios. Specifically, given a predefined ratio (e.g., 20\%) and a random seed, we randomly sample from the original training set. We obtain 5 training subsets by independently and repeatedly sampling five times from the original training set with 5 different random seeds. All methods are trained on these 5 subsets of the training set, and we report the average results on the test set. Figure~\ref{fig:training} shows results of CE, CE+AT, VIB, MEIB, and our CIFM against different sizes of training set with RoBERTa backbone. With a smaller ratio, the comparison methods struggle to capture complex patterns from the limited data, resulting in poor generalization on the test sets. Our CIFM achieves superior performance on all classification tasks against different ratios of the training set. This proves that CIFM can enhance the model's generalization ability, even under conditions of limited training data.

\subsection{Robustness Evaluation} 
We experiment to evaluate the model's robustness to noise samples under various optimization objectives during training. We adjust different strengths of random and adversarial perturbations on the test set. The random perturbations are from a multivariate Gaussian, and the adversarial perturbations are produced by a fast gradient method \cite{DBLP:conf/iclr/MiyatoDG17}. These perturbations are scaled by the $L_2$ norm and then applied to the embedding layer in the testing process. Following the empirical robust evaluation \cite{DBLP:conf/sp/Carlini017, DBLP:conf/acl/0001BWZH23}, we report the robust scores in terms of original evaluation metrics on noise samples generated from original test sets for each benchmark.

From Figure~\ref{fig:robust_random} and Figure~\ref{fig:robust_adv},
CIFM gains better robust scores over other objectives on all classification tasks. Compared to CE, CIFM achieves an average increase of \textbf{+3.2\%} and \textbf{+3.6\%} in robust scores under random and adversarial noise, respectively. It indicates that CIFM can safely discard the specific redundancy regions (i.e., region \textcircled{3} in Figure~\ref{fig:model}) that are negative to the target task. 
That is, CIFM learns noise-invariant features from input, which 
could be beneficial to noise scenes.

\subsection{Transferability Evaluation}
Unlike IB-based methods that learn representations by eliminating all target-irrelevant information, CIFM adversarially preserves noise-invariant but target-irrelevant features from the input, i.e., region \textcircled{2} in Figure~\ref{fig:model}. We believe that although the information of this region may not directly be relevant to the target, it can be valuable in learning more sufficient representations for the input data, accordingly to enhance the model's transferability. To prove this, we evaluate both linear and nonlinear transferability to new tasks. Two large-scale \textit{GoEmotions} and \textit{SentiEval} datasets are selected as the source dataset for pre-training. After pre-training, we freeze parameters of feature extractor (including parameters of an embedding layer and all hidden layers of the Transformer) and fine-tune a new classifier for each target task. 
We implement linear and nonlinear transferability evaluations by employing a linear hidden layer (Linear) and a convolutional neural network (CNN) as classifiers, respectively.

Table~\ref{tab:transfer_detailed1} summarizes the average scores over the other nine target tasks against two types of classifiers. Notably, the pre-trained CIFM consistently achieves the best average scores on new tasks by fine-tuning  linear and non-linear classifiers. The results demonstrate that CIFM can capture more sufficient representations from the input and the learned representations are more transferable. Besides, employing a CNN classifier tends to yield higher scores compared to a linear classifier. This proves nonlinear classifiers are instrumental in enhancing the model's adaptability across tasks.

\subsection{Representation Quality Evaluation} \label{sec:quality}

To evaluate the quality of the learned representations, we use two metrics to measure the sufficiency of both the input and target task.
Following \citet{DBLP:conf/icml/0001I20}, we choose the {uniformity}
metric to measure the preserved maximal information of hidden representations from the input. Following \citet{DBLP:conf/acl/0001BWZH23}, we apply the {adjusted rand index} (ARI) metric to assess the preserved maximal information of output representations for label structure. The lower the uniformity loss, the better sufficiency for the data. Conversely, a higher ARI score indicates a better sufficiency for the target.

Figure~\ref{fig:para} shows uniformity and ARI of  representations learned by various optimization objectives. CIFM not only extracts more uniform distributions on the hidden feature space, but also achieves better supervised clustering ability on the output representations. This implies that our CIFM exhibits more sufficient representations for the input and target, and the learned representations have good feature uniformity and sufficient predictive ability.

\section{Conclusion}
We propose an information-theoretic CIFM framework to 
extract noise-invariant sufficient representations for the input data and target task. Firstly, an IFM principle is proposed to learn more sufficient representations for the input and target. Besides, a CIM principle is designed to eliminate negative redundant features while preserve noise-invariant features from the input. Experiments on 13 language understanding benchmarks demonstrate that CIFM effectively improves the performance of PLMs for classification and regression.
Extensive experiments show that the learned representations are more sufficient, robust and transferable.

\section*{Limitations}
This paper introduces a new information-theoretic framework to 
extract sufficient representations for the input data and target task, as well as eliminates negative redundant features from the input. It promotes learned representations with good uniformity and sufficient predictive ability, thereby enhancing the generalization of pre-trained language models for the target task. However, the technique is studied only for different classification and regression tasks; its effectiveness in generation tasks remains to be further validated. Additionally, the principles proposed in this paper require more in-depth theoretical investigations.

\section*{Acknowledgements} 
This work was supported by the National Key Research and Development Program of China (No. 2022YFC3302102), and the Postdoctoral Fellowship Program of CPSF (No. GZC20232969). The authors thank the anonymous reviewers and the meta-reviewer for their helpful comments.

\bibliography{custom,acl}

\clearpage

\appendix

\begin{table*}[!ht]
\centering  
\small
\resizebox{0.85\linewidth}{!}{$
\begin{tabular}{llrrrrr} 
  \hline
  \multirow{1}{*}{\textbf{Dataset}} &
    \multirow{1}{*}{\textbf{Task}} 
    & \multicolumn{1}{c}{\# \textbf{Label}}  
    & \multicolumn{1}{c}{\# \textbf{Train}} 
    & \multicolumn{1}{c}{\# \textbf{Val}}  
    & \multicolumn{1}{c}{\# \textbf{Test}} 
    & \multicolumn{1}{c}{\# \textbf{Total}} 
    \\ 
    \hline
    \multicolumn{3}{l}{\textit{Classification Benchmarks}} \\ 
EmojiEval     &  Emoji prediction            & 20  &   45,000  &   5,000   &   50,000 & 100,000 \\ 
EmotionEval  & Social emotion detection          & 4   &   3,257   &   374     &   1,421 &  5,052 \\ 
HatEval   & Hate speech detection             & 2   &   9,000   &   1,000   &   2,970 & 12,970 \\ 
IronyEval     &   Irony detection            & 2   &   2,862   &   955     &   784 & 4,601\\ 
OffensEval &  Offensive language detection          & 2   &   11,916  &   1,324   &   860 & 14,100  \\ 
SentiEval &  Sentiment analysis              & 3   &   45,389  &   2,000   &   11,906 & 59.295  \\ 
StanceEval    &  Stance detection            & 3   &   2,620   &   294     &   1,249  & 4,163  \\ 
ISEAR     &  Emotion reaction prediction      & 7   &   3,066   &    767    &   3,833  & 7,666 \\
MELD      &  Conversational emotion recognition       &  7   &   9,989    &   1,109  &   2,610 & 13,708\\
GoEmotions &  Fine-grained emotion detection    & 28  &   36,308  &   4,548   &   4,591 & 45,447  \\
\hline
    \multicolumn{3}{l}{\textit{Regression Benchmarks}} \\ 
STS-B &  Semantic similarity prediction & - & 7,000 & 1,500 & 1,400 & 9,900 \\
CLAIRE & Plausible clarification ranking & - & 19,975 & 2,500 & 2,500 & 24,975 \\
EmoBank & Dimensional emotion analysis  & - & 8,062 & 1,000 & 1,000 & 10062   \\  \hline 
  \end{tabular}
  $}
  \caption{The statistics of all datasets and tasks. }
  \label{tab:datasets}
\end{table*}

\section*{Appendix Overview}
In the appendix, we will 
1) briefly illustrate related work, 
2) provide detailed experimental setups,
and 
3) report detailed experimental results.

\section{Related Work} \label{sec:rela}
Information-theoretic representation learning has become the workhorse of several impressive deep learning achievements, ranging from practical applications \cite{DBLP:conf/iclr/PereyraTCKH17, DBLP:conf/iclr/AlemiFD017,DBLP:conf/icml/BelghaziBROBHC18} to theoretical investigations \cite{xu2017information,DBLP:conf/colt/SteinkeZ20,DBLP:conf/icml/KawaguchiDJH23}.
The aim is typically to discover useful and informative latent representations in a principled and systematic manner.
In the context of the training principle, information-theoretic representation learning approaches can be broadly categorized into InfoMax-based and IB-based methods. 

In the field of machine learning, the information maximization (InfoMax) principle was first introduced by \citet{DBLP:journals/computer/Linsker88}. It suggests that a linear or nonlinear network can be viewed as an information channel and the goal is to transmit as much information as possible from the input data through the network. In recent
years, InfoMax 
has extensive applications in the field of 
self-supervised representation learning \cite{DBLP:conf/nips/ChenCDHSSA16,DBLP:journals/corr/abs-1807-03748,DBLP:conf/iclr/HjelmFLGBTB19, DBLP:conf/iclr/TschannenDRGL20,DBLP:conf/iclr/KongdYLDY20}. 
\citet{DBLP:conf/nips/ChenCDHSSA16} extend generative adversarial networks to learn disentangled representations by similarly using InfoMax.
\citet{DBLP:conf/iclr/TschannenDRGL20} study the usage of InfoMax in representation learning and shows the success depend on the inductive bias by the model architectures and mutual information (MI) estimators.
\citet{DBLP:conf/iclr/KongdYLDY20} explore several embedding models based on InfoMax, and introduce an objective that maximizes MI between the sentence representation and the n-grams.
In supervised scenarios, minimizing the standard cross-entropy is actually equivalent to maximizing the mutual information between the representations and the target task \cite{achille2018information,DBLP:journals/entropy/KolchinskyTW19,DBLP:conf/eccv/BoudiafRZGPPA20}. 
Moreover, \citet{DBLP:conf/iclr/PereyraTCKH17} 
add the negative entropy to the negative log-likelihood to penalize overconfident output distributions in a supervised manner.

The information bottleneck (IB) principle \cite{tishby1999information,tishby2015deep} aims to balance the trade-off between the compression of representation and the power of predicting. 
\citet{DBLP:conf/iclr/AlemiFD017} propose an efficient variational estimation method VIB based on the IB principle. 
\citet{DBLP:conf/iclr/PengKTAL19} use IB to constrain information flow in the discriminator for better adversarial learning.
Then, \citet{DBLP:conf/iclr/MahabadiBH21} use VIB to suppress irrelevant features when fine-tuning on low-resource language understanding tasks.
For tractable application of IB in a continuous setting, \citet{DBLP:conf/icml/BelghaziBROBHC18} propose a mutual information neural estimation method to estimate MI and design an IB principle based on the estimator.
\citet{DBLP:conf/cvpr/RagonesiVCM21} employ MINE-IB to learn unbiased representations.
\citet{DBLP:journals/entropy/Fischer20} and \citet{DBLP:conf/iclr/RameC21} introduce a conditional mutual information term to alleviate the over- and under-compression issue of IB.
\citet{DBLP:conf/cvpr/AnJC23} explicitly use the conditional entropy of the stochastic embedding as a confidence indicator and encourage the model to assign larger variance to more certain inputs.
Recently, \citet{DBLP:conf/aaai/0001WLZH24} propose a probabilistic coding framework to mitigate the over-compression issue of IB by simultaneously performing compression and prediction using a shared learnable network. They also introduced a regularization term to promote class-level uniformity for better task prediction.
Moreover, 
variational autoencoder (VAE) \cite{DBLP:journals/corr/KingmaW13} can be seen as a special case of an unsupervised VIB and have shown impressive results in self-supervised learning \cite{DBLP:conf/icml/Sohl-DicksteinW15,DBLP:conf/iclr/HigginsMPBGBML17,DBLP:conf/emnlp/0001HDZJMS22}.

\section{Detailed Experimental Setups}
\subsection{Downstream Datasets and Tasks} \label{sec:app:data}

We conduct experiments on 13 natural language understanding benchmarks, 
including 10 classification benchmarks (i.e., emoji prediction, hate speech detection, irony detection, offensive language detection, sentiment analysis, stance detection, and 4 categorical emotion analysis tasks from different domains), and 3 regression benchmarks (i.e., semantic similarity prediction, plausible clarifications ranking, and dimensional emotion analysis). 
The detailed statistics are shown in Table~\ref{tab:datasets}.

The descriptions of 10 classification benchmarks are listed as follows:
\textit{EmojiEval} \cite{DBLP:conf/semeval/BarbieriCRABBPS18} is designed for emoji prediction, which aims to predict its most likely emoji given a tweet. 
\textit{EmotionEval} \cite{DBLP:conf/semeval/MohammadBSK18} involves detecting the emotion evoked by a tweet and is based on the Affects in Tweets conducted during SemEval-2018.
Following \citet{DBLP:conf/emnlp/BarbieriCAN20}, the most common four emotions (i.e., anger, joy, sadness, and optimism) are selected as the label sets.
\textit{HatEval} \cite{DBLP:conf/semeval/BasileBFNPPRS19} stems from SemEval-2019 HatEval and is used to predict whether a tweet is hateful towards immigrants or women.
\textit{IronyEval} \cite{DBLP:conf/semeval/HeeLH18} is from SemEval-2018 Irony Detection and consists of identifying whether a tweet includes ironic intents or not.
\textit{OffensEval} \cite{DBLP:conf/semeval/ZampieriMNRFK19} is from SemEval-2019 OffensEval and involves predicting if a tweet contains any form of offensive language. 
\textit{SentiEval} \cite{DBLP:conf/semeval/RosenthalFN17} comes from SemEval-2017 and includes data from previous runs (2013,
2014, 2015, and 2016) of the same SemEval task. The goal is to determine if a tweet is positive, negative, or neutral. 
\textit{StanceEval} \cite{DBLP:conf/semeval/MohammadKSZC16} involves determining if the author's text has a favorable, neutral, or negative position towards a proposition or target. 
\textit{ISEAR} \cite{scherer1994evidence} is from International Survey On Emotion Antecedents And Reactions project and contains reports on seven emotions each by close to 3000 respondents in 37 countries on all 5 continents. 
It aims to predict the emotion reaction.
Due to the lack of a predefined split in the original corpus, we use the processed version by \citet{DBLP:conf/aaai/0001WLZH24} that randomly splits the dataset into train/valid/test sets in a ratio of 4:1:5 based on the label distribution.
\textit{MELD} \cite{DBLP:conf/acl/PoriaHMNCM19} contains multi-party conversation videos collected from Friends TV series.
It is used to detect emotions in each utterance.
The corpus contains many types of context, including dialogue, speaker, and multi-modal signals. Following \citet{DBLP:conf/aaai/0001WLZH24}, this paper only considers the context-free textual utterance to better evaluate sentence classification performance.
\textit{GoEmotions} \cite{DBLP:conf/acl/DemszkyMKCNR20} is a corpus of comments from Reddit, with human annotations to 27 emotion categories or neutral. It is used fine-grained emotion prediction. 
Following \citet{DBLP:conf/aaai/0001WLZH24}, nearly 16\% of multi-label data was removed from the source corpus to better evaluate the performance of multi-class classification.

The descriptions of 3 regression benchmarks are listed as follows:
\textit{STS-B} 
\cite{DBLP:conf/semeval/CerDALS17} is a collection of English sentence pairs drawn from news headlines, video and image captions, and natural language inference data. The semantic similarity prediction task is to predict the semantic textual similarity score from 0 (very dissimilar) to 5 (very similar) given each sentence pair.
\textit{CLAIRE} 
\cite{DBLP:conf/semeval/RothAS22} 
dataset consists of manually clarified how-to guides from wikiHow\footnote{https://www.wikihow.com/} with generated alternative clarifications and human plausibility judgements.
The goal of plausible clarifications ranking task is to predict the continuous plausibility score on a scale from 1 (very implausible) to 5 (very plausible) given the clarification and its context. 
In our experiments, a special token pair (i.e., \textless e\textgreater and \textless/e\textgreater) is introduced as the boundary of filler words.
\textit{Emobank} 
\cite{DBLP:conf/eacl/HahnB17} is a large-scale text corpus across 6 domains and 2 perspectives and manually annotated with continuous VAD scores. Each sentence has three scores representing VAD in the range of 1 to 5.

\begin{table*}[!ht]
\centering
\resizebox{\linewidth}{!}{$  
\begin{tabular}{c|l|ccccccccccccccc}
\hline 
\multicolumn{2}{c|}{\multirow{1}{*}{\textbf{Hyperparameter}}} 
& \multicolumn{1}{c}{{EmojiEval}} 
& \multicolumn{1}{c}{{EmotionEval}} 
& \multicolumn{1}{c}{{HatEval}} 
& \multicolumn{1}{c}{{IronyEval}} 
& \multicolumn{1}{c}{{OffensEval}} 
& \multicolumn{1}{c}{{SentiEval}} 
& \multicolumn{1}{c}{{StanceEval}} 
& \multicolumn{1}{c}{{ISEAR}} 
& \multicolumn{1}{c}{{MELD}} 
& \multicolumn{1}{c}{{GoEmotions}}  
& \multicolumn{1}{c}{{STS-B}}  
& \multicolumn{1}{c}{{CLAIRE}}  
& \multicolumn{1}{c}{{EmoBank}}  
\\ 
\hline 
\multicolumn{1}{l|}{\multirow{5}{*}{\rotatebox{90}{BERT}}}  
& Trade-off weight $\beta$  & 1 & 0.1 & 10 & 0.01 & 0.01 & 0.1 & 0.1 & 0.1 & 0.1 & 0.1 & - & - & - \\ 
& Temperature $\tau$   &  0.1 & 0.1 & 0.1 & 1 & 0.1 & 0.5 & 0.1 & 0.1 & 0.1 & 0.1 & - & - & -  \\
& Perturbation rate  & 1 & 1 & 1 & 1 & 0.1 & 1 & 1 & 1 & 1 & 1 & - & - & -  \\ 
& Perturbation radius $\epsilon$  & 0.1 & 1 & 0.1 & 5 & 5 & 1 & 1 & 5 & 1 & 0.1 & - & - & - \\ 
& Weight decay & 0 & 0.001 & 0.01 & 0.001 & 0 & 0 & 0.001 & 0 & 0.001 & 0 & - & - & - 
 \\ 
\hline 
\multicolumn{1}{l|}{\multirow{5}{*}{\rotatebox{90}{RoBERTa}}}  
& Trade-off weight $\beta$  & 0.01 & 1 & 10   & 1 & 1 & 0.01 & 0.1 & 0.1 & 1 & 0.1 & 0.001 & 0.01  & 0.01   \\ 
& Temperature $\tau$        & 0.1 & 0.5 & 0.1 & 1
& 1 & 0.1 & 0.5 & 0.1 & 1 & 0.1 & 0.1 & 1 & 1 \\
& Perturbation rate         & 1 & 0.1 & 1 & 0.1 & 1 & 1 & 1 & 1 & 1 & 1 & 1 & 0.1 & 0.1 \\ 
& Perturbation radius $\epsilon$  & 0.1 & 5 & 0.1 & 0.1
& 1 & 1 & 0.1 & 0.1 & 1 & 1  & 5 & 0.1 & 1\\ 
& Weight decay & 0 & 0 & 0.01 & 0.01 & 0.001 & 0 & 0 & 0 & 0 & 0 & 0 & 0  & 0
 \\ 
\hline 
\end{tabular}
$}
\caption{Hyperparameters of CIFM on 13 natural language understanding tasks.}
\label{tab:infoflow_para}
\end{table*}

\begin{table*}[!ht]
  \resizebox{0.99\linewidth}{!}{$
  \begin{tabular}{l|cccccccccc|c}
  \hline
  \multicolumn{1}{c|}{\multirow{1}{*}{{Methods}}} 
  & \multicolumn{1}{c}{{EmojiEval}} 
  & \multicolumn{1}{c}{{EmotionEval}} 
  & \multicolumn{1}{c}{{HatEval}} 
  & \multicolumn{1}{c}{{IronyEval}} 
  & \multicolumn{1}{c}{{OffensEval}} 
  & \multicolumn{1}{c}{{SentiEval}} 
  & \multicolumn{1}{c}{{StanceEval}} 
  & \multicolumn{1}{c}{{ISEAR}} 
  & \multicolumn{1}{c}{{MELD}} 
  & \multicolumn{1}{c|}{{GoEmotions}} 
  & \multicolumn{1}{c}{\textbf{Avg.}} 
  \\  
  \hline  
  \multicolumn{10}{l}{\it BERT backbone}  \\ \hline
\textbf{CIFM} & 
\textbf{24.28}\scriptsize{$\pm$0.39} & \textbf{77.74}\scriptsize{$\pm$1.07} & \textbf{59.22}\scriptsize{$\pm$1.75} & 
\textbf{65.51}\scriptsize{$\pm$1.61} & 
\textbf{80.67}\scriptsize{$\pm$0.99} & 
\textbf{70.66}\scriptsize{$\pm$0.28} & \textbf{67.81}\scriptsize{$\pm$0.97} & \textbf{71.51}\scriptsize{$\pm$0.51} & \textbf{41.72}\scriptsize{$\pm$0.80} & \textbf{48.42}\scriptsize{$\pm$0.74} & \textbf{60.75}
\\  
\ \ - w/o CIM & 
{24.05}\scriptsize{$\pm$0.35} & {77.30}\scriptsize{$\pm$0.62} & {55.69}\scriptsize{$\pm$2.11} & 
{60.89}\scriptsize{$\pm$3.11} & 
{80.35}\scriptsize{$\pm$1.19} & {70.12}\scriptsize{$\pm$0.73}  & {65.66}\scriptsize{$\pm$1.12} & {67.65}\scriptsize{$\pm$0.76} & {40.30}\scriptsize{$\pm$0.74} & {47.57}\scriptsize{$\pm$0.66} & 
58.96 \\ 
\ \ - w/o CIM \& IFM & {22.17}\scriptsize{$\pm$0.88} & {75.13}\scriptsize{$\pm$1.73} & {46.96}\scriptsize{$\pm$5.69} & {59.04}\scriptsize{$\pm$4.55} & {80.26}\scriptsize{$\pm$1.26} & {70.57}\scriptsize{$\pm$0.79} & {65.38}\scriptsize{$\pm$0.57} & {66.96}\scriptsize{$\pm$0.64} & {40.13}\scriptsize{$\pm$0.49} & {46.11}\scriptsize{$\pm$0.72} & 57.27
\\
\hline
  \multicolumn{10}{l}{\it RoBERTa backbone }  \\  \hline
\textbf{CIFM} & 
\textbf{32.32}\scriptsize{$\pm$0.87} & \textbf{79.63}\scriptsize{$\pm$0.57} & \textbf{59.55}\scriptsize{$\pm$0.91} & 
\textbf{63.02}\scriptsize{$\pm$3.81} 
& \textbf{80.96}\scriptsize{$\pm$0.55} & \textbf{72.93}\scriptsize{$\pm$0.27} & \textbf{69.01}\scriptsize{$\pm$0.80} & \textbf{71.66}\scriptsize{$\pm$0.68} & \textbf{43.99}\scriptsize{$\pm$1.10} & \textbf{49.51}\scriptsize{$\pm$0.31} & \textbf{62.26}
\\
\ \ - w/o CIM & 
{30.90}\scriptsize{$\pm$1.27} & {78.72}\scriptsize{$\pm$0.72} & {58.84}\scriptsize{$\pm$2.05} & 
{61.31}\scriptsize{$\pm$3.60} & 
{78.43}\scriptsize{$\pm$0.50} 
& {72.51}\scriptsize{$\pm$0.72} & {67.10}\scriptsize{$\pm$1.01} & {70.59}\scriptsize{$\pm$0.38} & {43.23}\scriptsize{$\pm$1.61} & {48.00}\scriptsize{$\pm$1.10} & 
60.96
\\
\ \ - w/o CIM \& IFM & 
{30.67}\scriptsize{$\pm$1.48} & {77.37}\scriptsize{$\pm$1.04} & {43.88}\scriptsize{$\pm$4.57} & 
{{47.01}\scriptsize{$\pm$7.05}} &
{80.29}\scriptsize{$\pm$0.35} & {72.74}\scriptsize{$\pm$0.40} & {67.60}\scriptsize{$\pm$0.74} & {70.42}\scriptsize{$\pm$0.47} & {41.30}\scriptsize{$\pm$1.67} & {47.96}\scriptsize{$\pm$0.73} & 
{57.92}
\\
\hline
\end{tabular}
$}
  \caption{Ablation results (\%) on 10 classification benchmarks. 
  We bolded best results for each task under BERT and RoBERTa backbones.
  }   \label{tab:abla2}
\end{table*}

\begin{table*}[!ht]
\centering
\small
\resizebox{0.86\linewidth}{!}{$
\begin{tabular}{l|ccccccc|c}
\hline 
\multicolumn{1}{c|}{\multirow{2}{*}{Methods}}    &  \multicolumn{2}{c}{STS-B} & \multicolumn{2}{c}{CLAIRE} & \multicolumn{3}{c|}{EmoBank} & \multirow{2}{*}{\bf Avg.}   \\  
& Spearman & Pearson & Spearman & Pearson & Pearson (v) & Pearson (a) & Pearson (d)  \\ 
\hline
\textbf{CIFM} &    \textbf{88.94}\tiny{$\pm$0.38} & \textbf{89.52}\tiny{$\pm$0.32} & \textbf{53.62}\tiny{$\pm$0.51} & \textbf{52.88}\tiny{$\pm$0.30} & \textbf{81.64}\tiny{$\pm$0.84} & 	\textbf{57.06}\tiny{$\pm$0.95} & 	\textbf{51.60}\tiny{$\pm$1.11} & \textbf{68.64} \\
\ \ - w/o CIM & {88.37}\tiny{$\pm$0.43} & {89.03}\tiny{$\pm$0.29} & {51.71}\tiny{$\pm$0.80} & {50.36}\tiny{$\pm$1.52} & {79.33}\tiny{$\pm$1.90} &	{55.01}\tiny{$\pm$1.82} &	{47.35}\tiny{$\pm$3.67} & 66.77 \\
\ \ - w/o CIM \& IFM & {88.44}\tiny{$\pm$0.42} & {89.10}\tiny{$\pm$0.37} & {51.45}\tiny{$\pm$1.00} & {50.82}\tiny{$\pm$1.36} &  {80.37}\tiny{$\pm$1.58} &	{55.83}\tiny{$\pm$0.68} &	{49.96}\tiny{$\pm$3.65} & 67.32 \\
\hline
\end{tabular}
$}
\caption{Ablation results (\%) on 3 regression benchmarks with RoBERTa backbone.
}
\label{tab:Regression_fine}
\end{table*}

\begin{table*}[!ht]
\centering
\resizebox{0.94\linewidth}{!}{$
\begin{tabular}{c|l|ccccccccc|c}
\hline 
\multicolumn{2}{c|}{\multirow{1}{*}{{Methods}}} & \multicolumn{9}{c|}{SentiEval $\rightarrow$ Others} & \multicolumn{1}{c}{\multirow{2}{*}{\bf Avg.}} \\ \cline{1-2}
Cls. &  \multicolumn{1}{c|}{Pre-train Obj.} & \multicolumn{1}{c}{{EmojiEval}} 
& \multicolumn{1}{c}{{EmotionEval}} 
& \multicolumn{1}{c}{{HatEval}} 
& \multicolumn{1}{c}{{IronyEval}} 
& \multicolumn{1}{c}{{OffensEval}} 
& \multicolumn{1}{c}{{StanceEval}} 
& \multicolumn{1}{c}{{ISEAR}} 
& \multicolumn{1}{c}{{MELD}} 
& \multicolumn{1}{c|}{{GoEmotions}} 
&   \\ 
\hline 
\multirow{5}{*}{Linear} 
& CE & 4.83 & 38.15 & 58.86 & 52.60 & 76.60 & 37.37 & 36.11 & 25.03 & 11.38 & 37.88 \\ 
& CE+AT & 4.89 & 37.81 & 57.54 & 51.62 & \bf 78.14 & 37.26 & 37.96 & 26.47 & 12.07 & 38.20 \\ 
& VIB & 3.81 & 42.14 & 57.38 & 50.51 & 76.08 & 36.69 & 34.54 & 23.90 & 8.55 & 37.07 \\ 
& MEIB & 3.51 & 37.33 & \bf 59.57 & 47.42 & 75.33 & 36.44 & 30.50 & 21.21 & 6.03 & 35.26 \\ 
& \bf CIFM & \bf 8.69 & \bf 43.04 & 56.42 & \bf 53.75 & 77.88 & \bf 39.59 & \bf 40.60 & \bf 28.98 & \bf 15.02 & {\bf 40.44} \\ 
\hline 
\multirow{5}{*}{CNN}
& CE & 24.37 & 73.83 & 51.49 & 64.82 & 80.07 & 62.67 & 60.52 & 41.71 & 36.66 & 55.13 \\ 
& CE+AT & 25.61 & 74.21 & 49.09 & \bf 66.62 & 80.43 & 62.70 & 61.17 & 41.34 & 37.41 & 55.40 \\ 
& VIB & 16.93 & 72.09 & \bf 54.66 & 64.11 & 79.51 & 59.69 & 57.22 & 40.51 & 32.95 & 53.07 \\ 
& MEIB & 13.94 & 70.81 & 54.32 & 62.25 & 78.78 & 57.44 & 52.48 & 38.54 & 29.87 & 50.94 \\ 
& \bf CIFM & \bf 26.43 & \bf 74.88 & 50.12 & 65.02 & \bf 80.64 & \bf 63.57 & \bf 62.81 & \bf 43.20 & \bf 39.23 & {\bf 56.21} \\ 
\hline
\end{tabular}
$}
\caption{Transferability results (\%) of \textit{SentiEval} $\rightarrow$ \textit{Others} with RoBERTa backbone.
Pre-train Obj. indicates the pre-training objective.
Cls. refers to the classifier when fine-tuning. 
Best results for linear (i.e., Linear) and nonlinear (i.e., CNN) classifiers are highlighted in bold.
}
\label{tab:transfer_detailed2}
\end{table*}

\subsection{Comparison Methods} \label{sec:comparison_method}
We compare against the 4 universal models (i.e., SVM, FastText, BiLSTM, and GPT-3.5) and 7 representative deep representation learning technologies (i.e., CE/MSE, CE+CP, CE/MSE+AT, CE+SCL, VIB, MINE-IB, and MEIB) with 2 different backbone models. VIB, MINE-IB, and MEIB belong to IB-based methods. The descriptions of these methods are listed as follows:

\textbf{SVM} \cite{cortes1995support} is a machine learning algorithm with a hinge loss that aims to find the best hyperplane to separate data points into different classes.
\textbf{FastText} \cite{DBLP:conf/eacl/GraveMJB17} is an efficient classification method with negative log-likelihood loss based on n-gram features and a hierarchical softmax.
\textbf{BiLSTM} is a bidirectional recurrent neural network \cite{hochreiter1997long}.
\textbf{GPT-3.5}\footnote{\url{https://chat.openai.com}} is an enhanced generative pre-trained transformer model based on text-davinci-003, optimized for chatting.

\textbf{CE/MSE} means a fine-tuned baseline with a cross-entropy (CE) loss for classification or a mean squared error (MSE) loss for regression.
\textbf{CE+CP} \cite{DBLP:conf/iclr/PereyraTCKH17} 
is an entropy regularization method that fits a deterministic network by optimizing an objective that combines the CE loss with a confidence penalty term. 
\textbf{MSE/CE+AT} \citep{DBLP:conf/iclr/MiyatoDG17} uses CE/MSE with classical adversarial training.
\textbf{CE+SCL} \cite{gunel2020supervised}
combines CE and supervised contrastive learning (SCL) \cite{khosla2020supervised}.
SCL allows for multiple positives per anchor, thus adapting contrastive learning to the fully supervised setting.
\textbf{VIB} \cite{DBLP:conf/iclr/AlemiFD017,DBLP:conf/iclr/MahabadiBH21}
is an efficient variational estimation method based on IB \cite{tishby2015deep}. 
\textbf{MINE-IB} \cite{DBLP:conf/icml/BelghaziBROBHC18}
is a neural estimation method based on IB with a continuous setting.
\textbf{MEIB}  
\cite{DBLP:conf/cvpr/AnJC23} 
is a variational approach to stochastic embedding in which maximum entropy acts as the bottleneck.

\subsection{Implementation Details} \label{sec:para}

All methods are conducted with the epoch number of $20$, total batch size of $128$, and maximum token length of $128$. The maximum patience of early stopping is set to 5 epochs. The dropout rate is set to 0.2.
For the default InfoNCE estimator used in IFM, the trade-off parameter $\beta$  is searched from $\{0.01, 0.1, 1, 10\}$ for classification and $\{0.001, 0.01, 0.1\}$ for regression, and the temperature $\tau$ is searched from $\{0.1, 0.5, 1\}$.
For the adversarial estimator used in CIM, the $L_q$ norm constraint is $L_2$, the perturbation rate is searched from $\{0.1, 1\}$, and the perturbation radius is searched from $\{0.1, 1, 5\}$.  
Table~\ref{tab:infoflow_para} shows the best parameters of our method with RoBERTa and BERT backbones.

For the compared GPT-3.5, we present the zero-shot results of the GPT-3.5-turbo snapshot from June 13th 2023 based on specific inputs, including task descriptions, instructions, and evaluation texts.

\section{Supplementary Results} 
\subsection{Details of Ablation Studies} \label{sec:results_ablated}
Table~\ref{tab:abla2} and Table~\ref{tab:Regression_fine} report detailed ablation results on classification and regression tasks, respectively.

\subsection{Details of Transferability Evaluation}  
\label{tab:transferability}
Table~\ref{tab:transfer_detailed2} shows detailed transferability results of \textit{SentiEval} $\rightarrow$ \textit{Others}.

\end{document}